\documentclass[10pt,twocolumn,letterpaper]{article}

\usepackage{cvpr}              
\usepackage[table]{xcolor}
\usepackage{algorithm}
\usepackage{algorithmic}
\usepackage{newfloat}
\usepackage{listings}

\usepackage{amsmath}
\usepackage{tipa}
\usepackage{pifont}
\usepackage{amsfonts}
\usepackage{multirow} 
\usepackage{graphicx} 
\usepackage{booktabs} 
\definecolor{cvprblue}{rgb}{0.21,0.49,0.74}
\usepackage[pagebackref,breaklinks,colorlinks,allcolors=cvprblue]{hyperref}

\def\paperID{46282} 
\def\confName{CVPR}
\def\confYear{2026}

\begin{document}
\title{HM-Talker: Hybrid Motion Modeling for High-Fidelity Talking Head Synthesis}

\author{Shiyu Liu$^1$\quad Kui Jiang$^1$\thanks{Corresponding author.}\quad Junjun Jiang$^1$\quad
Xianming Liu$^1$ \quad \\ Xiaocheng Feng$^1$\quad Fei Ma$^2$\quad Hongxun Yao$^1$\quad Qi Tian$^2$ \\
$^1$Harbin Institute of Technology University\quad \\ $^2$Guangdong Laboratory of Artificial Intelligence and Digital Economy (sz)\quad  
}

\maketitle

\begin{abstract}
\noindent Audio-driven talking head generation faces a fundamental trade-off between personalization and generalization, limiting its practical application. Implicit models often achieve generalization at the cost of structural incoherence, resulting in unstable head motion and inaccurate lip synchronization. While explicit methods incorporate geometric and anatomical priors such as 3D Morphable Models (3DMMs), which parameterize facial geometry, or Action Units (AUs), which code facial muscle movements—they tend to produce overly neutral expressions or suffer from limited generalization. To resolve this conflict, we present HM-Talker, an audio-driven talking head framework that synergistically integrates explicit articulatory cues with implicit prosodic features to characterize identity-specific dynamics while enabling audio-driven generalization. Its distinctive features can be summarized as: i) the Cross-Modal Mapping Module (CMMM) that extracts a comprehensive vocabulary of motion cues from audio and video, and ii) the Hybrid Motion Modeling Module (HMMM) that employs a Stochastic Feature Pairing (SFP) strategy to dynamically merge paired implicit and explicit features for motion synthesis. This design facilitates an iterative optimization of the lower face motion, alternating between identity-specific and identity-agnostic (audio-only) objectives. Extensive experiments demonstrate that HM-Talker outperforms state-of-the-art methods in both visual realism and lip-sync accuracy across diverse settings.

\end{abstract}
    
\section{Introduction}
\label{sec:intro}
\begin{figure}[!ht]
    \centering
    \includegraphics[width=1\linewidth]{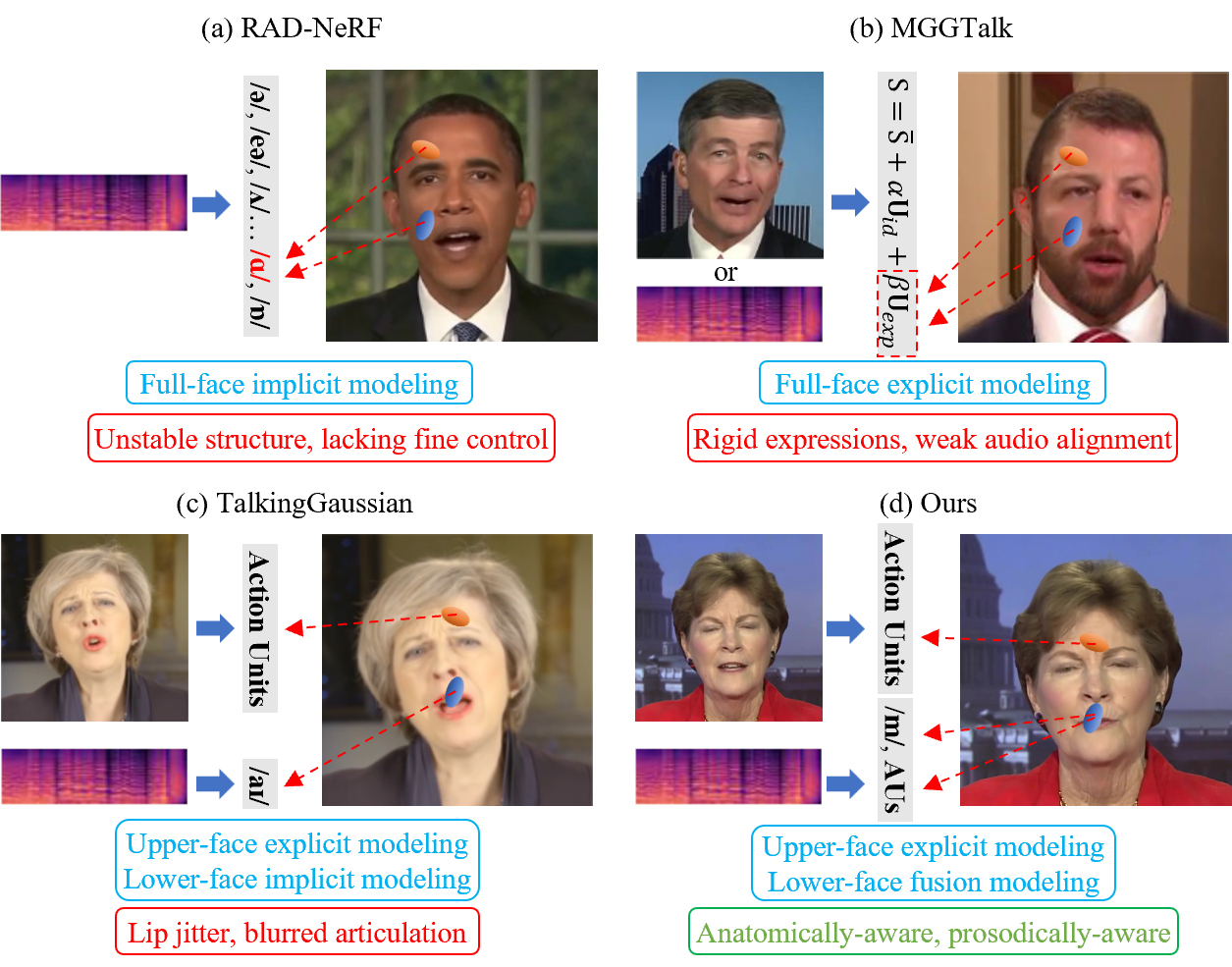}
    \vspace{-6mm}
    \caption{
    The Dilemma in Motion Modeling. (a) Purely implicit models lack structural control, leading to artifacts. (b) Purely explicit models struggle with prosody, causing rigid motion. (c) A spatial compromise inherits implicit flaws in the mouth region, resulting in lip jitter. (d) Our feature-level fusion model directly addresses this trade-off, achieving both anatomical and prosodic awareness.
    }
    \label{fig:intro}
    \vspace{-6mm}
\end{figure}
The pursuit of photorealistic audio-driven talking heads remains a central challenge in computer vision and graphics. 
Early methods~\cite{prajwal2020lip,zhong2023identity,zhang2023dinet} map acoustic features directly to lip movements using generative networks or employ primitive 3D representations~\cite{mildenhall2021nerf,kerbl20233d} for motion modeling. 
Despite recent remarkable progress~\cite{peng2024synctalk,cho2024gaussiantalker,li2025talkinggaussian}, state-of-the-art methods still exhibit temporal instability and local artifacts such as lip jitter. 

A milestone work, InsTaG~\cite{li2025instaglearningpersonalized3d} achieves impressive performance by decomposing the identity-specific motion into universal priors and personalized adaptation fields, revealing that high-fidelity talking-head synthesis could benefit from combining generalizable motion structures with individual expressive dynamics.
Consequently, a fundamental tension exists: an ideal model must learn a hybrid motion field that captures identity-specific facial features and speaking style while generalizing robustly to unseen, identity-agnostic audio. 

Balancing personalization and generalization, however, is non-trivial. Implicit models~\cite{guo2021ad,tang2022real,li2023efficient} learn a direct audio-to-motion mapping. Despite various technical refinements such as enhanced feature encoding~\cite{tang2022real,li2023efficient}, these methods inherently lack structural constraints, often producing unstable or distorted facial dynamics (\cref{fig:intro}(a)).
Conversely, explicit methods that incorporate geometric or anatomical priors introduce other limitations. 
Priors derived from statistical multi-person models, such as 3D Morphable Models~\cite{10.1145/311535.311556,blanz2023morphable} (3DMMs), improve generalizable animation~\cite{gong2025monoculargeneralizablegaussiantalking,chu2024generalizableanimatablegaussianhead} but often sacrifice person-specific expressiveness (\cref{fig:intro}(b)). In contrast, priors extracted directly from a target video, like per-frame Action Units~\cite{ekman1978facial} (AUs), provide rich stylistic cues but risk overfitting and poor generalization. 
Even recent hybrid methods, such as TalkingGaussian~\cite{li2025talkinggaussian}, only partially resolve this issue. These approaches often use explicit priors to stabilize the upper face, but rely on an unconstrained implicit model for the highly articulated lower face. By omitting structural guidance where it is most critical, this compromise leads to a loss of per-frame precision, which accumulates over time as temporal artifacts like lip jitter (\cref{fig:intro}(c)). 

This analysis yields our key insight: photorealistic talking head generation requires leveraging person-specific (geometric and anatomical) priors while ensuring generalization to unseen (cross-identity) audio. To this end, we propose HM-Talker, a hybrid motion modeling framework for universal audio-driven facial animation. 
Unlike prior methods that rely solely on image-derived explicit features or audio-derived implicit features for lower face motion, HM-Talker harmonizes both merits to enable anatomical grounding-based reconstruction (prosodic features) while enabling audio-driven generalization. 

Our framework achieves this balance through two synergistic components. First, the \textbf{Cross-Modal Mapping Module (CMMM)} constructs a comprehensive motion vocabulary by extracting implicit prosodic features from audio and explicit articulatory features from the reference video, while 
projecting audio features into the visual articulatory space to align with image-based representation via an Audio-to-Visual Mapper. This enables grounding-based reconstruction even under audio-driven conditions.
Second, the \textbf{Hybrid Motion Modeling Module (HMMM)} employs 
a Stochastic Feature Pairing (SFP) strategy to adjust input features. Specifically, this training scheme dynamically alternates between fusing implicit audio features with (1) video-derived explicit priors to enforce personalization, and (2) audio-predicted explicit features to alleviate identity-dependent biases in explicit motion and enhance cross-identity generalization under audio-driven conditions. 

Our main contributions are threefold:
\begin{itemize}
\item We introduce HM-Talker, a novel paradigm for audio-driven talking head generation by strategically reconciling the fundamental trade-off between personalization and generalization.
\item We propose the Cross-Modal Mapping Module (CMMM), which prepares a comprehensive vocabulary of motion cues from both audio and video sources, enabling the model to synthesize anatomically plausible facial movements.
\item We design the Hybrid Motion Modeling Module (HMMM) with a Stochastic Feature Pairing strategy, where dynamical optimization alternately reinforces personalization and generalization to learn a robust and adaptable facial motion synthesizer.
\end{itemize}

\section{Method}

\begin{figure*}[!ht]
    \centering
    \includegraphics[width=0.95\linewidth]{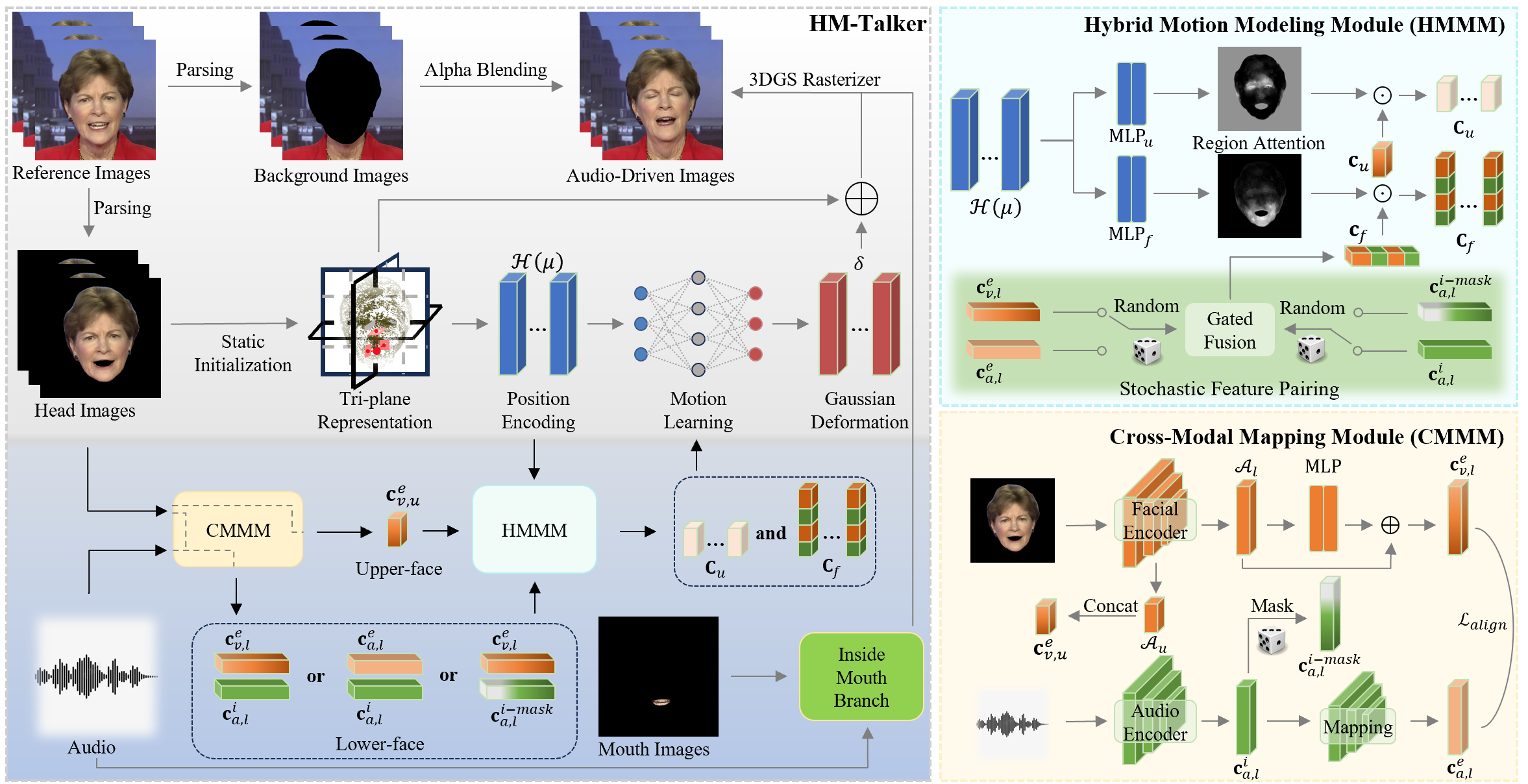}
    \vspace{-3mm}
    \caption{
    \textbf{Pipeline of the HM-Talker.} Given audio input and a series of reference head images, a static Gaussian field is initialized. A Tri-plane Encoder extracts positional encoding $\mathcal{H}(\mu)$ from this field. Concurrently, the audio and reference image are processed by our \textbf{Cross-Modal Mapping Module (CMMM)} to generate a vocabulary of motion cues, including implicit features ($c^i_{a,l}$) and various explicit features ($c^e_{v,u}, c^e_{v,l}, c^e_{a,l}$). These cues are then fed into the \textbf{Hybrid Motion Modeling Module (HMMM)}. Within HMMM, a stochastic fusion mechanism generates the final lower face control vector $\mathbf{C}_f$ and upper face control vector $\mathbf{C}_u$. These vectors, conditioned on the positional encoding $\mathcal{H}(\mu)$, are used by the \textbf{Deformation Network} to predict the complete Gaussian deformations $\delta$. Finally, the deformed Gaussians are rendered, alpha-blended with the inside-mouth output to generate the final talking head video.
    }
    \label{fig:pipeline}
\vspace{-4mm}
\end{figure*}

\subsection{Preliminary: Deformable Gaussian Avatars}
\label{sec:preliminary}

Our framework builds upon the deformable 3D Gaussian Avatars paradigm, specifically adopting \textit{TalkingGaussian}~\cite{li2025talkinggaussian} as the foundational backbone. This approach offers high-fidelity rendering and an explicit decomposition of the talking head into a static identity representation and a dynamic motion field.

\noindent\textbf{Canonical Representation.}
A subject’s identity and geometry are encoded by a set of \textit{Persistent 3D Gaussians}, parameterized by $\theta = \{\mu, s, q, \alpha, f\}$, which represents 3D position, scale, rotation, opacity, and spherical harmonics coefficients, respectively. The Gaussian centers $\mu$ in this canonical space collectively form a deformable template capturing the subject’s unique facial structure.

\noindent\textbf{Dynamic Motion Representation.}
Facial animation is driven by a motion field, which predicts a per-Gaussian 3D displacement $\delta = \{\Delta \mu, \Delta s, \Delta q\}$ 
from the canonical template. 
This displacement is conditioned on the Gaussian’s canonical position $\mu$ and a condition vector $\mathbf{C}$:
\begin{equation}
\delta = \text{MLP}(\mathcal{H}(\mu) \oplus \mathbf{C}),
\end{equation}
where $\mathcal{H}(\cdot)$ is tri-plane hash encoding~\cite{li2023efficient} 
for high-frequency spatial localization, and $\oplus$ indicates feature concatenation. The condition vector $\mathbf{C}$ incorporates features from the driving signal (\emph{e.g.}, audio) alongside explicit parameters such as expression coefficients. 
The objective is to learn a mapping to $\delta$ so that the resulting motion field 
produces realistic and subject-specific facial animations. 

\noindent\textbf{Limitation of TalkingGaussian.}
In \textit{TalkingGaussian}, the motion field $\mathbf{C}$ is derived via a spatially partitioned scheme that combines implicit audio features with explicit expression parameters. Specifically, the upper face is governed primarily by explicit priors, whereas the lower face is driven predominantly by audio cues. 
Although this design effectively disentangles facial regions, it compromises stability. Specifically, the absence of explicit structural guidance for the lower face often results in motion jitter and imprecise lip synchronization. To resolve this instability, we introduce a hybrid motion formulation by integrating implicit and explicit cues for the lower face reconstruction while the dynamical optimization alternately reinforces personalization and generalization.

\subsection{Overview of HM-Talker}
\label{sec:overview}

To overcome these limitations, we propose \textbf{HM-Talker}, a hybrid motion modeling framework designed to learn a motion field that balances generalization accuracy and personalized modeling. 
As illustrated in \cref{fig:pipeline}, our design focuses on improving motion learning specifically for the lower face region—where speech-related dynamics are most complex—while preserving explicit, prior-driven guidance for the upper face. 
A key feature of HM-Talker is its \textbf{prior-flexible} design: while we instantiate explicit motion representations using Action Units (AUs) for concreteness, 
the framework is readily adaptable to other motion priors, such as 3D Morphable Models (3DMMs) and BlendShapes~\cite{2014Practice}.

HM-Talker comprises two synergistic components. The \textbf{Cross-Modal Mapping Module (CMMM)} first extracts explicit facial representations from video, distinguishing between the upper ($c^e_{v,u}$) and lower ($c^e_{v,l}$) regions.  
It also derives implicit motion features from audio ($c^i_{a,l}$) and projects them into a visual articulatory space ($c^e_{a,l}$) to establish cross-modal alignment. The \textbf{Hybrid Motion Modeling Module (HMMM)} subsequently fuses these implicit and explicit cues via a gated attention. Crucially, during training, we employ a stochastic pairing of feature sets to promote identity-specific personalization while simultaneously enhancing generalization to unseen speakers and audios.

HM-Talker is trained in a multi-stage paradigm, 
encompassing static identity initialization, motion learning, and fine-tuning. The overall objective is defined as:
\begin{equation}
\mathcal{L}_{\text{total}} = \mathcal{L}_{1} + \lambda_1 \mathcal{L}_{\text{D-SSIM}} + \lambda_2 \mathcal{L}_{\text{LPIPS}} + \lambda_3 \mathcal{L}_{\text{align}},
\end{equation}
where $\mathcal{L}_{1}$, $\mathcal{L}_{\text{D-SSIM}}$, and $\mathcal{L}_{\text{LPIPS}}$ ensure pixel-level and perceptual fidelity. The alignment term is depicted as
\begin{equation}
\mathcal{L}_{\text{align}} = \mathcal{L}_1(\mathbf{c}_{a,l}^{e}, \mathbf{c}_{v,l}^{e}),
\label{eq:align}
\end{equation}
enforcing consistency between audio-derived and visual articulatory embeddings, thereby grounding the audio-driven motion in the subject’s physical expression.
For ease of exposition, we 
focus our description on the \textit{Face Branch}; the \textit{Inside Mouth Branch} 
follows an analogous hybrid-driven strategy as the lower facial motion pathway.

\subsection{Cross-Modal Mapping Module (CMMM)}
\label{sec:cmmm}

\textbf{CMMM} establishes a motion vocabulary by projecting implicit audio features into explicit articulatory representations. This explicit cue space serves as a reliable guide for subsequent motion modeling.

\noindent\textbf{Target Space: Video-Derived Explicit Representation.}
The target space is constructed from the ground-truth video data, providing a high-fidelity supervisory signal and motion cues for learning personalized motion. We extract frame-wise Facial Action Units (AUs) and divide them into upper face ($\mathcal{A}_u$) and lower face ($\mathcal{A}_l$) subsets. For the upper face, we form the representation $\mathbf{c}_{v,u}^{e}$ via a straightforward concatenation of AUs, which is sufficient to capture the relatively rigid and less articulated dynamics of this region. The upper face representation $\mathbf{c}_{v,u}^{e}$ is directly passed to the motion field decoder without modification, following the original TalkingGaussian framework.

For the lower face AUs, which govern complex articulatory motions, we employ a residual MLP to enhance expressiveness:
\begin{equation}
    \mathbf{c}_{v,l}^{e} = \mathrm{MLP}(\mathcal{A}_l) \oplus \mathcal{A}_l \in \mathbb{R}^{32}.
\end{equation}
The output dimension of $32$ is chosen to match that of the audio-derived features, facilitating subsequent fusion. 
The residual connection ensures that the original anatomical information is preserved, while the non-linear branch captures complex correlations between AUs. This explicit vector $\mathbf{c}_{v,l}^{e}$ encodes the subject’s personalized articulatory style, 
serving as the primary supervision for the audio-to-motion projection and as a direct motion cue in our hybrid model.

\noindent\textbf{Source Space: Audio-Derived Implicit Representation.}
The source space is derived from the input audio using a pre-trained Audio-Visual Encoder (AVE)~\cite{peng2024synctalk}, which outputs a prosody-rich feature $a \in \mathbb{R}^{512}$. These features are then processed by a two-stage network designed to distill articulatory information. First, \textit{AudioNet} acts as a feature extractor, performing hierarchical temporal compression to produce compact embeddings. Subsequently, \textit{AudioAttNet} 
refines these embeddings using an attention mechanism to focus on 
perceptually salient temporal regions, yielding the final implicit feature $\mathbf{c}_{a,l}^{i} \in \mathbb{R}^{32}$. The implicit feature $\mathbf{c}_{a,l}^{i}$ captures prosodic and temporal audio cues, thereby complementing the explicit visual features to generate accurate and personalized lower face motion. (Architectural details for AudioNet and AudioAttNet are provided in the Appendix).

\noindent\textbf{Cross-Modal Projection.}
To project audio-derived features into the explicit articulatory space, we introduce Audio-to-Visual Mapper (A2VM), a lightweight MLP defined as 
\begin{equation}
    \mathbf{c}_{a,l}^{e} = \mathrm{A2VM}(\mathbf{c}_{a,l}^{i}).
\end{equation}
This projection is supervised by the alignment loss $\mathcal{L}_{\text{align}}$ (\cref{eq:align}), which encourages the predicted $\mathbf{c}_{a,l}^{e}$ to match the video-derived target $\mathbf{c}_{v,l}^{e}$. By learning this mapping, the CMMM produces lower face features that integrate audio-derived motion cues with explicit visual priors, subsequently serving as inputs to the HMMM.

\subsection{Hybrid Motion Modeling Module (HMMM)}
\label{sec:hmmm}

\textbf{HMMM} is designed to resolve the fundamental trade-off between personalization and generalization in talking-head synthesis. It consumes the vocabulary of motion cues from the CMMM and integrates our key technical contribution, the \textbf{Stochastic Feature Pairing (SFP)} strategy, with a dynamically-gated fusion mechanism.

\noindent\textbf{Dual-Objective training in SFP.}
SFP operates by stochastically alternating between distinct feature-pairing paths during training, thereby balancing two primary objectives: capturing personalized motion and generalizing to unseen audio. At each iteration, one of the following paths is selected:

\begin{itemize}
    \item \textbf{Path 1 ($\mathcal{P}_{\text{style}}$) - Supervised Personalization:} Pairs the implicit audio feature $\mathbf{c}_{a,l}^{i}$ with the video-derived explicit representation $\mathbf{c}_{v,l}^{e}$. 
    Conditioned on the referenced motion cues, this path anchors the learned motion field to the subject’s unique articulatory style. 
    \item \textbf{Path 2 ($\mathcal{P}_{\text{gen}}$) - Audio-Driven Generalization:} Pairs $\mathbf{c}_{a,l}^{i}$ with the audio-predicted explicit feature $\mathbf{c}_{a,l}^{e}$. This path, deprived of ground-truth video priors, compels the model to generate accurate motion from audio alone, ensuring a generalizable audio-to-motion mapping.
    \item \textbf{Path 3 ($\mathcal{P}_{\text{robust}}$) - Auxiliary Robustness:} Pairs a masked audio feature $\mathbf{c}_{a,l}^{i\text{-mask}} = \mathcal{M}_a \cdot \mathbf{c}_{a,l}^{i}$ with $\mathbf{c}_{v,l}^{e}$. $\mathcal{M}_a$ is an element-wise random mask sampled uniformly (\emph{e.g.}, $[0.1,0.3]$), scaling 
    feature dimension to improve robustness and prevent over-reliance on the audio stream.  

\end{itemize}

Unlike conventional multi-task learning that computes all losses simultaneously, 
our stochastic path selection 
is highly memory-efficient, as only one path is active per iteration.
During inference, only the generalization path ($\mathcal{P}_{\text{gen}}$) is used, ensuring fully deterministic, audio-driven motion synthesis.

\noindent\textbf{Dynamically-Gated Fusion.}
The explicit and implicit features selected by SFP are fused through a learnable gating mechanism:
\begin{equation}
    \mathbf{c}_f = \alpha \odot \mathbf{c}_{\cdot,l}^{e} + (1 - \alpha) \odot \mathbf{c}_{a,l}^{i*},
\end{equation}
where the gating weight is predicted via a lightweight MLP with a sigmoid activation:
\begin{equation}
    \alpha = \mathrm{sigmoid}(\mathrm{MLP_g}(\mathbf{c}_{a,l}^{i*} \oplus \mathbf{c}_{\cdot,l}^{e})),
\end{equation}
where $\mathbf{c}_{a,l}^{i*}=\{\mathbf{c}_{a,l}^{i},\mathbf{c}_{a,l}^{i-mask}\}$ and $\mathbf{c}_{\cdot,l}^{e}=\{\mathbf{c}_{a,l}^{e},\mathbf{c}_{v,l}^{e}\}$.
This mechanism allows the model to adaptively balance the influence of structured explicit cues and prosodic implicit cues on a frame-by-frame basis.

\noindent\textbf{Final Deformation Prediction.}
The fused lower face feature $\mathbf{c}_f$ and the upper face feature $\mathbf{c}_{v,u}^e$ are modulated by spatially-aware attention maps~\cite{guo2022beyond} to enable fine-grained, region-specific control. 
These maps are predicted by two lightweight MLPs ($\text{MLP}_f, \text{MLP}_u$) from the positional encoding $\mathcal{H}(\mu)$, depicted as 
\begin{equation}
    \mathbf{C}_f = \mathbf{c}_f \odot \text{MLP}_\text{f}(\mathcal{H}(\mu)),
    \mathbf{C}_u = \mathbf{c}_{v,u}^e \odot \text{MLP}_\text{u}(\mathcal{H}(\mu)).
\end{equation}
The final per-Gaussian deformation $\delta_{\text{face}}$ is predicted by a deformation MLP that integrates all conditioned features: 
\begin{equation}
\delta_{\text{face}} = \text{MLP}(\mathcal{H}(\mu) \oplus \mathbf{C}_u \oplus \mathbf{C}_f).
\end{equation}
This hierarchical and spatially-attentive conditioning ensures the generated motion is both holistically coherent and preserves the detailed, personalized articulatory style learned. 

\section{Experiments}
\subsection{Experimental Settings}

\noindent\textbf{Dataset.} 
Following established protocols in~\cite{ye2023geneface, li2023efficient}, we evaluate our method on five publicly available portrait videos to ensure fair and impartial comparisons. The dataset comprises three male subjects (``Lieu", ``Jae-in", and ``Obama") and two female subjects (``May" and ``Shaheen"), with an average duration of 7,637 frames captured at 25 FPS. All recordings maintain portrait-centered composition, predominantly at 512$\times$512 resolution except for 450$\times$450 resolutions for ``Obama" and ``Jae-in". 
 
\noindent\textbf{Baselines for Comparison.} 
Our comparative analysis encompasses three distinct categories of contemporary approaches: 2D generative models (IP-LAP~\cite{zhong2023identity}, TalkLip~\cite{wang2023seeing}, DINet~\cite{zhang2023dinet}), Neural Radiance Field (NeRF) based methods (AD-NeRF~\cite{guo2021ad}, RAD-NeRF~\cite{tang2022real}, ER-NeRF~\cite{li2023efficient}, SyncTalk~\cite{peng2024synctalk}), and 3D Gaussian Splatting (3DGS) based techniques (GaussianTalker~\cite{cho2024gaussiantalker}, TalkingGaussian~\cite{li2025talkinggaussian}). To further ensure a fair comparison, we re-implemented the TalkingGaussian baseline using our data preprocessing pipeline like SyncTalk's, which is denoted as TalkingGaussian in our results.

\noindent\textbf{Static Image Quality.} We employ Peak Signal-to-Noise Ratio (PSNR), Structural Similarity Index (SSIM)~\cite{wang2004image}, and Learned Perceptual Image Patch Similarity (LPIPS)~\cite{zhang2018unreasonable} metrics for evaluation.

\noindent\textbf{Dynamic Motion Quality.} Lip synchronization is evaluated using SyncNet~\cite{chung2017lip, chung2017out}, which provides a Confidence Score (Sync-C) and an Error Distance (Sync-D). We also measure the Landmark Distance (LMD)~\cite{chen2018lip} between generated and reference facial expressions. To further analyze the articulatory precision, we use Action Units (AUs)~\cite{prince2015facial} extracted via OpenFace~\cite{baltrusaitis2018openface} and report the error for the lower facial region (AUE-L) and upper facial region (AUE-U).

\noindent\textbf{Efficiency.} We report the total Training Time (in hours) for model convergence on a single subject and the inference speed in Frames Per Second (FPS) on a 512$\times$512 sequence.


\noindent\textbf{Implementation Details.} 
For each subject, we first train the Face Branch and Inside Mouth Branch in parallel for 50,000 iterations. During this stage, the Face Branch is driven by hybrid motion features selected according to the HMMM's three-path strategy, with path proportions set to $\mathcal{P}_{\text{gen}}:\mathcal{P}_{\text{robust}}:\mathcal{P}_{\text{style}} = 4{:}4{:}2$ and the audio masking ratio $\mathcal{M}_a$ sampled uniformly from 0.1 to 0.3.  Both branches are then jointly fine-tuned for an additional 15,000 iterations to stabilize articulation consistency. We use Adam~\cite{kingma2014adam} and AdamW~\cite{loshchilov2018adamw} optimizers with a learning rate of $5\mathrm{e}{-4}$, and set loss weights to $\lambda_1 = 0.2$, $\lambda_2 = 0.5$, and $\lambda_3 = 1\mathrm{e}{-3}$. All experiments are conducted on a single NVIDIA RTX 3090 GPU. During inference, upper-face motion is driven by image-based explicit features, while lower-face and inner-mouth regions are generated from audio-only inputs.


 \begin{table*}[!ht]

\begin{center}

\resizebox{0.95\linewidth}{!}{
\setlength{\tabcolsep}{3.1mm}

\begin{tabular}{@{}c@{\hspace{12pt}}>{\hspace{-8pt}}l|ccc|ccc|cc} 
\toprule
\addlinespace[0mm]
 \rowcolor{gray!50} 
\multicolumn{2}{l|}{ \multirow{2}{*}{ }}
 & \multicolumn{3}{c|}{  \textbf{Rendering Quality}} & \multicolumn{3}{c|}{ \textbf{Motion Quality}} & \multicolumn{2}{c}{ \textbf{Efficiency}} \\ 
 \addlinespace[-0.1mm]
 \rowcolor{gray!50}
  \multicolumn{2}{l|}{ \multirow{-2}{*}{  \textbf{Methods}}}    & \textbf{PSNR} $\uparrow$ &  \textbf{LPIPS} $\downarrow$ & \textbf{SSIM} $\uparrow$  & \textbf{LMD} $\downarrow$ & \textbf{AUE-(L/U)} $\downarrow$ & \textbf{Sync-C} $\uparrow$ & \textbf{Time} & \textbf{FPS} \\
  \addlinespace[-0.6mm]
    \midrule
    \addlinespace[0mm]
    \multirow{4}{*}{\rotatebox[origin=c]{90}{NeRF}} & AD-NeRF~\cite{guo2021ad} & 30.07 & 0.1042 & 0.9689 & 2.998 & 1.01/0.97 & 6.053 & 18.7h & 0.11  \\
                                                  & RAD-NeRF~\cite{tang2022real} & 31.95 & 0.0620 & 0.9660 & 2.847 & 0.74/0.76 & 5.742 & 5.3h & 28.7  \\
                                                  & ER-NeRF~\cite{li2023efficient} & 32.47 & 0.0395 & 0.9658 & 2.639 & 0.62/0.54 & 6.531 & 2.1h & 31.2 \\
                                                  & SyncTalk~\cite{peng2024synctalk} & \underline{34.51} & \underline{0.0221} & \underline{0.9959} & \underline{2.607} & \underline{0.55}/0.29 & \underline{7.502}  & 2.0h & 52 \\
    \addlinespace[-0.6mm]
    \midrule
    \addlinespace[0mm]
     & GaussianTalker~\cite{cho2024gaussiantalker} & 32.69 & 0.0442 & 0.9952 & 2.726 & 0.67/0.59 & 6.234 & 3.2h  & 95 \\
                                                  & TalkingGaussian~\cite{li2025talkinggaussian}
                                                  & 32.48 & 0.0309 & 0.9950 & 2.616 & 0.60/0.28 & 6.246 & \textbf{0.5h}  & 108 \\
                                                  & TalkingGaussian*~\cite{li2025talkinggaussian}
                                                  & 33.75 & 0.0273 & 0.9963 & 2.699 & 0.59/\underline{0.26} & 6.451 & \underline{0.55h}  & \textbf{139} \\
                                                  \addlinespace[-0.6mm]
    \cmidrule(l){2-10} 
    \addlinespace[-1.0mm]
    
    \multirow{-4}{*}{\rotatebox[origin=c]{90}{3DGS}}                                              & \cellcolor{gray!15}HM-Talker (Ours) & \cellcolor{gray!15}\textbf{35.15} & \cellcolor{gray!15}\textbf{0.0207} &\cellcolor{gray!15}\textbf{0.9971} & \cellcolor{gray!15}\textbf{2.514} &\cellcolor{gray!15}\textbf{0.53}/\textbf{0.22} &\cellcolor{gray!15}\textbf{7.807} &\cellcolor{gray!15} 0.6h & \cellcolor{gray!15}\underline{120}  \\
    \addlinespace[-0.6mm]
    \bottomrule
    \vspace{-6mm}
\end{tabular}}

\end{center}
\vspace{-4mm}
\caption{Comparison with 3D-aware methods on self-reconstruction. We attain leading performance across the majority of metrics when compared to methods based on NeRF or 3DGS. The best and second-best results are indicated in bold and with underlines, respectively. \textbf{TalkingGaussian*} denotes our re-implementation of the method using the same data preprocessing pipeline as SyncTalk and our approach.}
\label{tab:sr1}
\vspace{-6mm}
\end{table*}

\subsection{Comparison with SOTA}

\begin{table}[!b]
\setlength\tabcolsep{5pt}
\begin{center}
\vspace{-4mm}
\resizebox{0.9\linewidth}{!}{
\begin{tabular}{@{}c@{\hspace{12pt}}>{\hspace{-8pt}}l|ccc} 

\toprule
\addlinespace[0mm]
\rowcolor{gray!50} 
\multicolumn{2}{l|}{\multirow{2}{*}{}}
& \multicolumn{3}{c}{\textbf{Motion Quality}}  \\ 
\addlinespace[-0.1mm]
\rowcolor{gray!50} 
\multicolumn{2}{l|}{\multirow{-2}{*}{\textbf{Methods}}}       & \textbf{LMD} $\downarrow$ & \textbf{AUE-(L/U)} $\downarrow$ & \textbf{Sync-C} $\uparrow$  \\
\addlinespace[-0.6mm]
    \midrule
    \addlinespace[0mm]
    \multirow{3}{*}{\rotatebox[origin=c]{90}{GAN}}
    & Wav2Lip~\cite{chung2017lip}  & 2.948 & 0.70/- & \textbf{8.755}  \\
                                                  & IP-LAP~\cite{zhong2023identity}  & 3.161 & 1.00/- & 7.040  \\
                                                  & DINet~\cite{zhang2023dinet}   & 3.230 & 1.09/- & 7.455  \\
                                                  & TalkLip~\cite{wang2023seeing}   & 3.285 & 0.82/- & 6.657  \\
    \addlinespace[-0.6mm]
    \midrule
    \addlinespace[0mm]
    ~    &\cellcolor{gray!15}HM-Talker (Ours)  &  \cellcolor{gray!15}\textbf{2.514} & \cellcolor{gray!15}\textbf{0.53}/\textbf{0.22} & \cellcolor{gray!15} 7.807  \\
    \addlinespace[-0.6mm]
    \bottomrule
\end{tabular}}
\end{center}
\vspace{-6mm}
\caption{Motion quality comparison with 2D generative methods. HM-Talker achieves significantly better results across all metrics. The `-' for upper-face AU error (AUE-U) indicates that a reliable measurement could not be obtained, as 2D methods often generate motion only for the mouth region, while keeping the rest of the face from the original frame unchanged.}
\label{tab:sr2}
\end{table}
\noindent\textbf{Self-Reconstruction.}
We evaluate HM-Talker under a 10:1 train–validation split across all datasets. As shown in \cref{tab:sr1,tab:sr2}, our method achieves leading performance in both visual quality and motion fidelity. It outperforms SyncTalk (34.51 dB) and TalkingGaussian* (33.75 dB) with a PSNR of \textbf{35.15 dB}, while achieving the lowest motion errors (LMD =\ \textbf{2.514}, AUE-L =\ \textbf{0.53}) and the highest synchronization confidence (Sync-C =\ \textbf{7.807}).
These results support the effectiveness of explicitly learning a \textbf{personalized motion field} for achieving more stable and accurate facial dynamics. HM-Talker’s hybrid implicit–explicit framework captures individual articulatory patterns more faithfully, helping to mitigate lip jitter and improve synchronization.
Although 2D generative models specialize in synchronization, HM-Talker surpasses top-performing baselines such as Wav2Lip on most motion metrics, while preserving superior 3D realism. Furthermore, it maintains real-time rendering speed (120 FPS) and fast convergence (0.6 h), matching the efficiency of 3DGS-based pipelines.

\begin{figure}[!ht]
    \centering
    \includegraphics[width=0.9\linewidth]{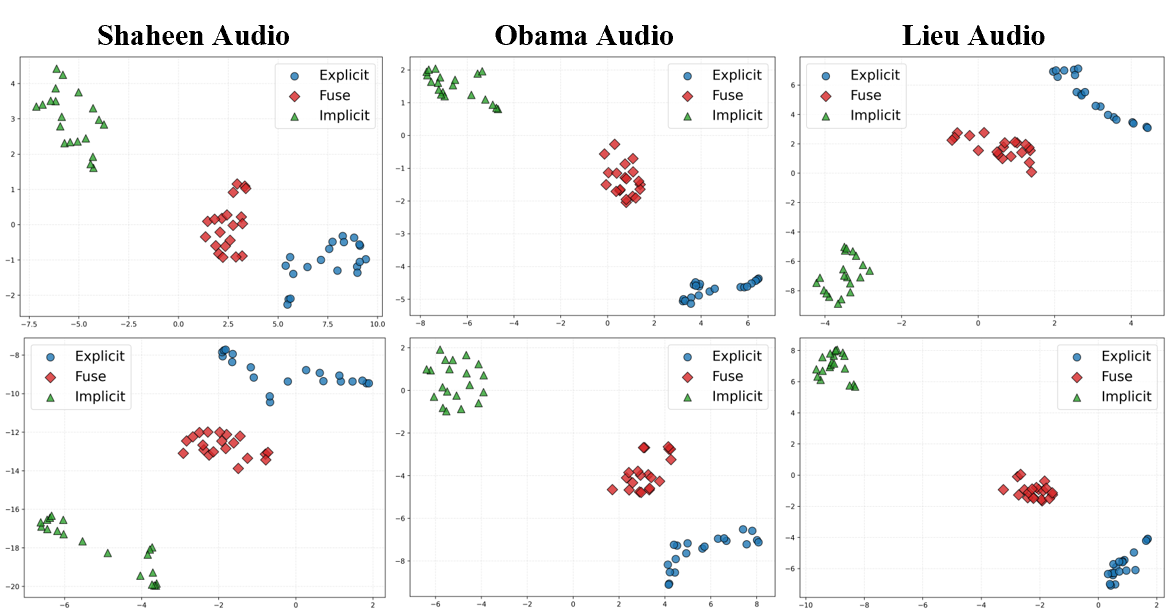}
    \vspace{-4mm}
    \caption{
 t-SNE visualization showing modality fusion. Implicit audio (green) and explicit visual (blue) features form distinct clusters, while the fused representations (red) create a unified manifold, indicating effective cross-modal alignment.}
    \label{fig:tSNE}
\vspace{-6mm}
\end{figure}

\noindent\textbf{Generalization to Unseen Audio and Speakers.}  
To assess generalization, we drive models trained solely on the ``May'' dataset using out-of-domain audio from unseen speakers—``Shaheen'' (gender-matched) and ``Lieu'' (gender-mismatched). As shown in~\cref{tab:ls}, HM-Talker achieves superior lip-synchronization accuracy across these settings, outperforming SyncTalk by a noticeable margin. 
These results indicate that the Hybrid Motion Modeling Module effectively disentangles phonetic content from speaker-specific acoustics, 
thereby facilitating a robust content-to-style mapping. Furthermore, the Stochastic Feature Pairing (SFP) strategy acts as a regularizer, explicitly encouraging generalization beyond the training speaker's voice. 
The t-SNE visualization in~\cref{fig:tSNE} reveals that while implicit audio and explicit visual features reside in distinct subspaces, 
their fused representations form a continuous manifold bridging the two modalities, indicating effective cross-modal alignment and robust feature fusion.

\begin{table}[!b]
\begin{center}
\vspace{-2mm}

\resizebox{\linewidth}{!}{

\begin{tabular}{l|cccc}
\toprule
\addlinespace[0mm]
\rowcolor{gray!50} 
 
             & \multicolumn{2}{c}{\textbf{``Shaheen" Audio}} & \multicolumn{2}{c}{\textbf{``Lieu" Audio}} \\ 
             \addlinespace[-0.8mm]
             \cmidrule(l){2-5} 
             \addlinespace[-1.0mm]
\rowcolor{gray!50} 
      \multirow{-2}{*}{\textbf{Methods}}         & \textbf{Sync-D} $\downarrow$   & \textbf{Sync-C} $\uparrow$  & \textbf{Sync-D} $\downarrow$   & \textbf{Sync-C} $\uparrow$  \\
\addlinespace[-0.6mm]
             \midrule
                                        \addlinespace[0mm]
DINet~\cite{zhang2023dinet}          
                                        & \underline{8.201}         & \underline{7.295}               
                                        & 8.226         & 6.470          \\ 
IP-LAP~\cite{zhong2023identity}          
                                        & 9.819             & 5.316      
                                        & 9.392             & 5.077       \\ 
TalkLip~\cite{wang2023seeing}          
                                        & 9.553             & 5.488      
                                        & 11.679             & 3.151       \\ 
                                        \addlinespace[-0.6mm]
                                        \midrule
                                        \addlinespace[0mm]

RAD-NeRF~\cite{tang2022real}         
                                        & 12.012       & 3.054                                  
                                        & 12.044       & 2.449                             \\
ER-NeRF~\cite{li2023efficient}      
                                        & 9.775             & 5.529           
                                        & 10.017             & 4.782         \\ 

SyncTalk~\cite{peng2024synctalk}        
                                        &  8.903        & 6.350                   
                                        & 7.508         & 7.780                \\

GaussianTalker~\cite{cho2024gaussiantalker}      
                                        & 8.926       & 6.576             
                                        & 10.943       & 4.198           \\ 
                                        
TalkingGaussian~\cite{li2025talkinggaussian}     
                                        & 11.450       & 3.179             
                                        & 9.849       & 5.039            \\ 
TalkingGaussian*~\cite{li2025talkinggaussian}     
                                        & 8.283       &  6.768            
                                        & \underline{7.439}       & \underline{7.803}            \\ 
                                        \addlinespace[-0.6mm]
                                        \midrule
                                        \addlinespace[0mm]
                                        \rowcolor{gray!15}
HM-Talker (Ours)     
                                        & \textbf{7.590}    & \textbf{7.972}     
                                        & \textbf{7.292}    & \textbf{7.994}            \\ 
                                        \addlinespace[-0.6mm]
                                        \bottomrule

\end{tabular}
 }
\end{center}
\vspace{-6mm}
\caption{Cross-identity audio driving comparison. HM-Talker delivers superior lip-sync accuracy when driving the ``May" avatar with unseen audios (``Shaheen", ``Lieu"). 
}
\label{tab:ls}
\end{table}

\begin{figure*}[!ht]
    \centering
    \includegraphics[width=0.94\linewidth]{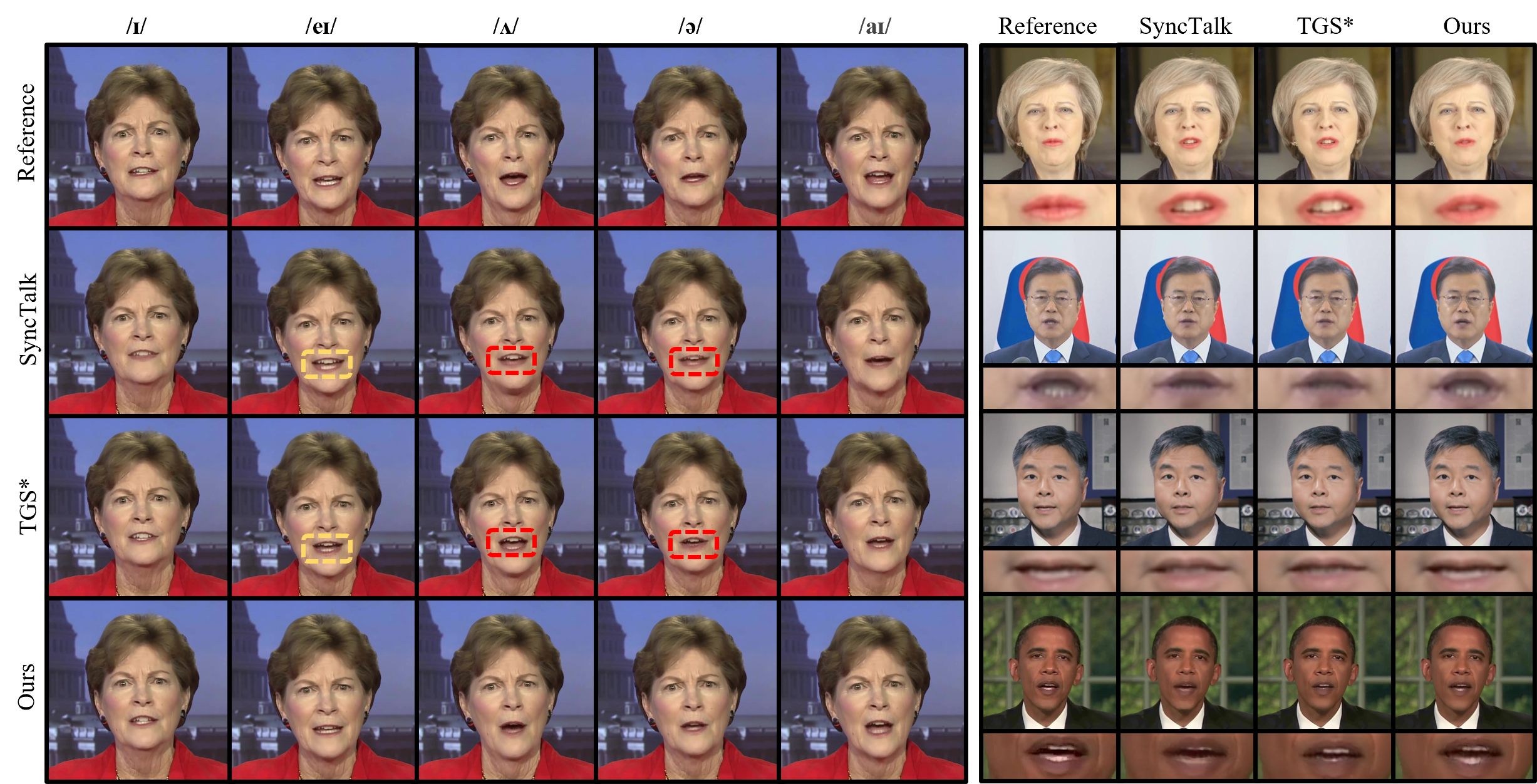}
    \vspace{-2mm}
    \caption{\textbf{Qualitative Comparison.} Visual results on challenging phonemes. Our method achieves the most consistent phoneme–viseme alignment and preserves intra-oral realism. ``TGS*'' denotes ``TalkingGaussian*''. Zoom in for better visualization.}
    \label{fig:IQ}
    \vspace{-6mm}
    
\end{figure*}

\noindent\textbf{Qualitative Comparison.}
We conduct qualitative, frame-level comparisons 
against state-of-the-art temporal modeling approaches: SyncTalk, TalkingGaussian*, and our HM-Talker. Key video frames corresponding to target phonemes are selected to critically assess phoneme-viseme alignment. As illustrated in \cref{fig:IQ}, our approach generates the most visually consistent results 
with reference frames across various phoneme categories. For instance, during the articulation of wide-mouth phonemes (\emph{e.g.}, \textipa{/eI/}) or subtle ones (\emph{e.g.}, \textipa{/@/}), 
our model maintains precise lip closure and shape, whereas competing methods exhibit noticeable misalignments, as indicated by the red boxes. 
For articulations such as \textipa{/\textturnv/}, while baseline methods capture a broadly similar mouth aperture, 
our method reconstructs more fine-grained intra-oral details (highlighted in yellow), 
achieving superior perceptual realism. These observations demonstrate the efficacy of applying our hybrid motion modeling to the complete articulatory system. By jointly modeling external facial expressions and internal oral structures with a unified, hybrid-driven strategy, our framework ensures cohesive and synchronized  movement across all components, resulting in realistic talking head synthesis.

\begin{figure}[!ht]
    \centering
    \includegraphics[width=0.9\linewidth]{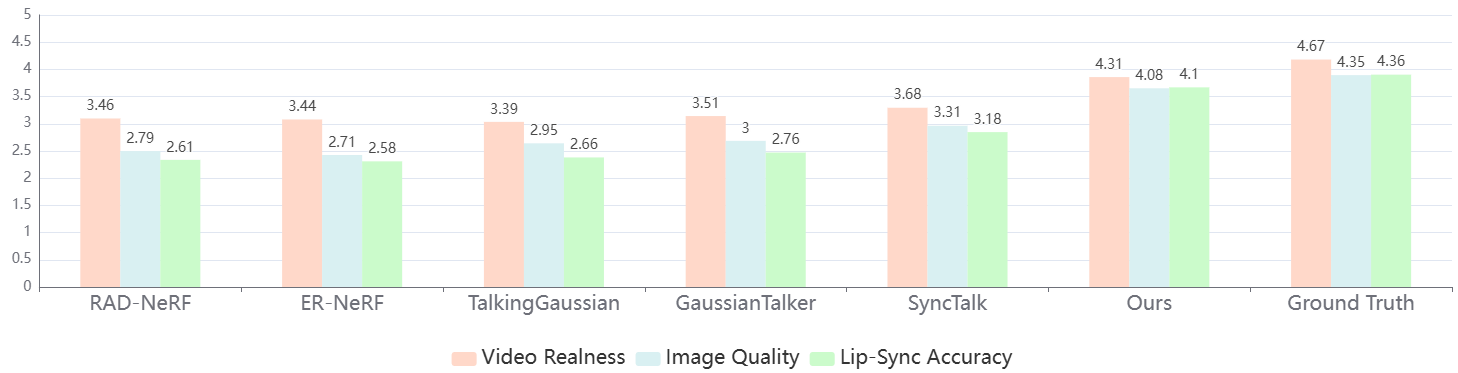}
    \vspace{-4mm}
    \caption{User Study. Mean Opinion Scores (MOS) on a 5-point scale. Our method achieves the best scores than all evaluated baselines across three metrics, 
    closer to the ground truth.}
    \label{fig:UserStudy}
    \vspace{-4mm}
\end{figure}

\noindent\textbf{User Study.}
We conduct a user study with 30 non-expert participants evaluating 35 videos (5 identities $\times$7 methods including ground truth), each 20 seconds in duration. Participants rate Video Realness, Image Quality, and Lip-Sync Accuracy using a 5-point scale. Our method consistently outperforms all competitors across all metrics (\cref{fig:UserStudy}).
In particular, our method achieves 4.31 score in Video Realness and 4.08 in Image Quality, surpassing the second-best method by margins of 17\% and 23\%, respectively.
Moreover, our approach gains 4.10 score (Lip-Sync Accuracy), significantly narrowing the gap to the ground truth (4.36).
\begin{figure}[!b]
    \centering
    \vspace{-6mm}
    \includegraphics[width=0.9\linewidth]{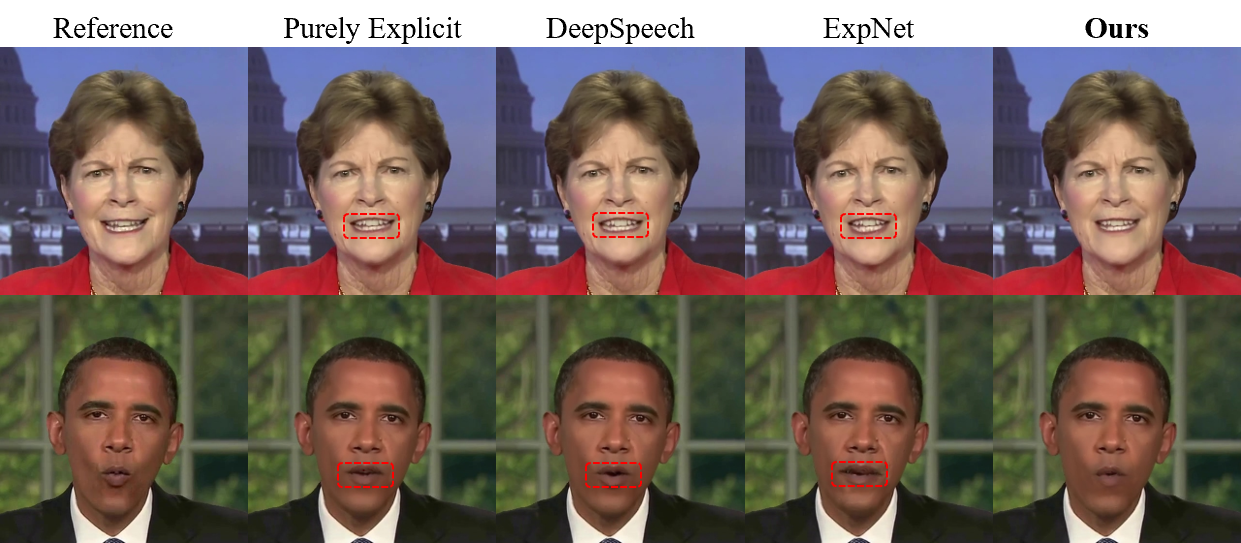}
    \vspace{-4mm}
    \caption{Visual results of the ablation study. The incomplete models, using generic encoders or lacking hybrid fusion, exhibit clear artifacts such as inaccurate lip shapes and motion stiffness.}
    \label{fig:ablation_visual}
\vspace{-4mm}
\end{figure}

\subsection{Ablation Studies on Self-Reconstruction}
\label{sec:ablation_studies}

We perform comprehensive ablations to analyze the individual influence of fusion strategy and component configurations on the reconstruction stability and fidelity. 

\noindent\textbf{Fusion Strategy.}
We first ablate the fusion mechanism, which is responsible for integrating explicit structural priors with implicit prosodic cues to form a stable and expressive \textbf{personalized motion field}. We compare four variants: 
(a) \textit{Purely Implicit}: setting $\alpha=0$ during training, similar to the lower-face configuration in TalkingGaussian*;
(b) \textit{Purely Explicit}: setting $\alpha=1$ during training, making the model depend solely on the explicit motion predicted by CMMM;
(c) \textit{MLP Fusion}, directly concatenating the two features, followed by an MLP;   
(d) \textit{Gated Fusion (ours)}, our proposed adaptive fusion. 

The results reveal a clear hierarchy (\cref{tab:ablation_all}). The unimodal baselines (a, b) perform poorly, confirming the necessity for hybrid modeling. While MLP Fusion (c) improves consistency but still lacks adaptive control for optimal performance. In contrast, our Gated Fusion (d) achieves the best performance, with an LMD of \textbf{2.514} and Sync-C of \textbf{7.807}. This result validates that an adaptive gating mechanism is crucial for effectively combining personalized structural style with dynamic prosodic information. 

\noindent\textbf{Component Robustness.}
We next evaluate the robustness and contribution of individual components by ablating the following:
(1) replacing the Action Unit prior with 3DMM or BlendShape~\cite{2014Practice} coefficients;  
(2) replacing the learnable Audio-to-Visual Mapper with a pretrained ExpNet encoder from SadTalker~\cite{zhang2023sadtalkerlearningrealistic3d};  
(3) using a different audio encoder (DeepSpeech~\cite{hannun2014deepspeech});   
(4) replacing the explicit motion stream with Gaussian noise.

\begin{figure}[b!]
    \centering
    \includegraphics[width=0.9\linewidth]{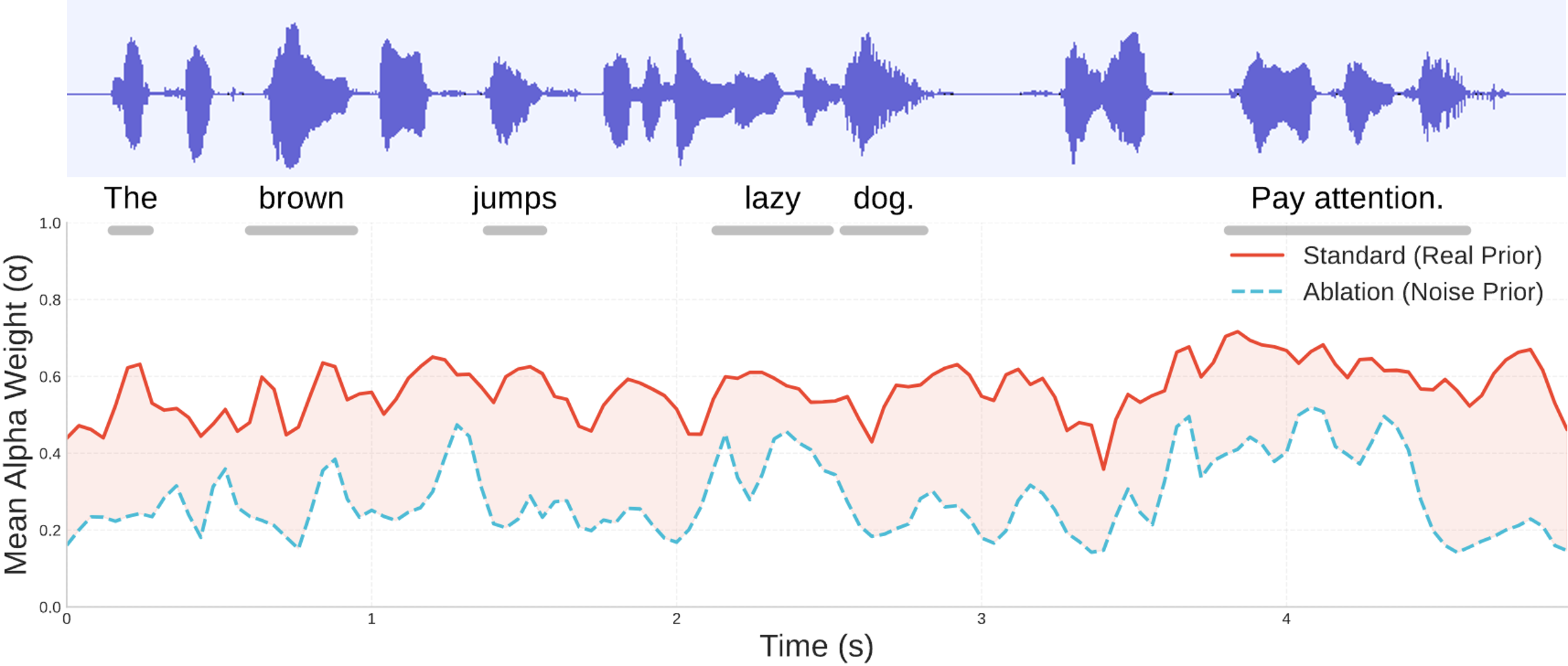}
    \vspace{-6mm}
    \caption{
    Analysis of HMMM's adaptive gating. Learned weights $\alpha$ under standard (\textcolor{red}{red solid}) and noise-prior (\textcolor{blue}{blue dashed}) settings show that the model down-weights unreliable priors and loses fine-grained responses without structural cues.
    }\vspace{-5mm}
    \label{fig:alpha_analysis}
\end{figure}

Results reveal three key observations (\cref{tab:ablation_all}). Replacing the AU prior with 3DMM or BlendShape yields comparable results, showing that HMMM is \textbf{prior-flexible}.  Replacing our A2VM with the fixed ExpNet encoder causes a notable drop in lip-sync accuracy, underscoring the benefit of our identity-specific, learnable audio-to-visual mapping. 
When using the less suitable DeepSpeech features, the fusion gate $\alpha$ converges toward~1, and performance drops. This indicates that HMMM intelligently down-weights unreliable or noisy audio streams, relying more heavily on the trustworthy explicit pathway.
Replacing the explicit stream with noise causes $\alpha$ to drop to 0.2 (compared to ~0.6 in the default setting). As illustrated in \cref{fig:ablation_visual}, such incomplete variants often produce inaccurate lip shapes or stiff motion. Furthermore, as shown in \cref{fig:alpha_analysis}, $\alpha$ exhibits dynamic, word-dependent fluctuations: for ``The" and ``jumps", it rises then falls in our model but stays flat or inverted under noise; for ``lazy", the trend reverses. These patterns confirm that HMMM adaptively balances the two streams based on the reliability of the input cues.

\begin{table}[htb!]
\centering
\resizebox{\linewidth}{!}{
\begin{tabular}{l|cccc}
\toprule
\addlinespace[0mm]
\rowcolor{gray!50}
\textbf{Setting} & \textbf{PSNR}$\uparrow$ & \textbf{AUE-(L/U)}$\downarrow$ & \textbf{Sync-C}$\uparrow$ & \textbf{LMD}$\downarrow$ \\
\addlinespace[-0.6mm]
\midrule
\multicolumn{5}{l}{\textit{\textbf{Analysis of Fusion Strategy}}} \\
(a) Purely Implicit ($\alpha=0$) & 33.82 & 0.59/0.26 & 6.485 & 2.686 \\
(b) Purely Explicit ($\alpha=1$) & 34.30 & 0.58/0.28 & 6.895 & 2.681 \\
(c) MLP Fusion & 35.05 & 0.54/0.31& 7.770 & 2.527 \\
\rowcolor{gray!15} (d) Gated Fusion (Ours) & \textbf{35.15} & 0.53/\textbf{0.22} & \textbf{7.807} & \textbf{2.514} \\
\addlinespace[-0.6mm]
\midrule
\multicolumn{5}{l}{\textit{\textbf{Analysis of Component Choices}}} \\
HM-Talker w/ 3DMM & 35.12 & \textbf{0.52}/0.26 & 7.679 & 2.534 \\
HM-Talker w/ BlendShape & 35.11 & 0.53/0.25 & 7.731 & 2.520 \\
HM-Talker w/ Noise & 34.10 & 0.58/0.25 & 6.592 & 2.692 \\
HM-Talker w/ ExpNet & 34.26 & 0.85/0.25 & 6.404 & 3.181 \\
HM-Talker w/ DeepSpeech & 34.85 & 0.72/0.22 & 6.230 & 2.718 \\
\addlinespace[-0.6mm]
\bottomrule
\end{tabular}
}
\vspace{-3mm}
\caption{Ablation results under different fusion strategies and component configurations. 
}
\vspace{-4mm}
\label{tab:ablation_all}
\end{table}

\subsection{Ablation Studies on Audio Generalization}
To verify our proposed generalization mechanisms, we perform ablation studies on two components: the Audio-to-Visual Mapper (A2VM) and the Stochastic Feature Pairing (SFP) strategy. Experiments are conducted under increasing distribution shifts, including cross-identity and cross-lingual scenarios. The results are tabulated in \cref{tab:cross}.

\noindent\textbf{Ablation Setup.} We define four variants to isolate the effects of each component: 
(i) ablating \textbf{A2VM} by removing the module and its alignment loss, leaving SFP’s generalization path fully implicit; 
(ii) ablating \textbf{SFP} by disabling the generalization-focused paths ($P_{\text{gen}}$, $P_{\text{robust}}$) and training only with the personalization objective ($P_{\text{style}}$); 
(iii) ablating both components, reducing the model to a fully implicit baseline; 
(iv) the full model with all components enabled.

\noindent\textbf{Analysis of Stochastic Feature Pairing (SFP).} 
Ablation studies confirm the indispensable role of SFP in enabling model generalization. Removing SFP results in a severe performance drop in both cross-identity (Sync-C: 8.663 $\rightarrow$ 3.727) and cross-lingual (6.501 $\rightarrow$ 1.871) settings. This decline is attributable to severe over-fitting on the source speaker's identity and linguistic patterns when the model is trained solely on the personalization objective. These results collectively underscore that SFP's multi-objective, stochastic training strategy is essential for learning a generalizable audio-to-motion mapping.


\noindent\textbf{Analysis of Audio-to-Visual Mapper (A2VM).} 
The role of A2VM is nuanced and task-dependent. In in-distribution data (S-to-S), removing A2VM slightly decreases performance (Sync-C: 10.269 $\rightarrow$ 9.683), confirming its role as a structural regularizer that enhances articulatory precision. Conversely, in the cross-lingual task (M-to-S), A2VM ablation yields an improvement in Sync-D (lower is better) yet a degradation in Sync-C (6.501 $\rightarrow$ 6.183). This indicates a trade-off: relying on a purely implicit mapping improves temporal consistency (as reflected in Sync-D) but at the expense of phonetic accuracy (Sync-C). Nevertheless, even under unseen language conditions, A2VM provides crucial structural guidance that enforces more precise viseme shapes. In summary, these findings indicate that cross-domain generalization is primarily enabled by the robust audio encoder, while A2VM serves as a precision-enhancing module, the effectiveness of which is modulated by the degree of domain shift.

\noindent\textbf{Summary.} Our analysis reveals that SFP provides a robust, domain-agnostic foundation for generalization, while A2VM offers a powerful but potentially domain-specific refinement. The full model, combining both, achieves the best overall trade-off between high-fidelity reconstruction and robust generalization across diverse speech conditions.

\begin{table}[]
\centering
\resizebox{\linewidth}{!}{
    \begin{tabular}{l|cc|ccc}
    
    \toprule
    \addlinespace[0mm]
    \rowcolor{gray!50}
    \textbf{Setting} &\textbf{A2VM}  & \textbf{SFP} & \textbf{Sync-C}$\uparrow$ & \textbf{Sync-D} $\downarrow$   \\ 
    \addlinespace[-0.6mm]
    \midrule
    \addlinespace[0mm]
   \multirow{4}{*}{S(English)-to-S(English)}  &\cellcolor{gray!15}\ding{51} &\cellcolor{gray!15}\ding{51} &\cellcolor{gray!15}\textbf{10.269} &\cellcolor{gray!15}\textbf{5.860}   \\
    & \ding{51} & \ding{55} & 10.066 & 6.125  \\ 
    & \ding{55} & \ding{51} & 9.683 & 6.557  \\ 
    & \ding{55} & \ding{55} & 9.248 & 6.612  \\ 
    \addlinespace[-0.6mm]
    \midrule
    \addlinespace[0mm]
   \multirow{4}{*}{L(English)-to-S(English)}  &\cellcolor{gray!15}\ding{51} 
   &\cellcolor{gray!15}\ding{51}
   &\cellcolor{gray!15}\textbf{8.663} &\cellcolor{gray!15}\textbf{6.589}   \\
    & \ding{51} 
    & \ding{55}
    & 3.727 & 11.729  \\ 
    & \ding{55}
    & \ding{51} 
    & 8.462 & 7.245  \\ 
    &  \ding{55}
    &  \ding{55}
    & 8.326 & 6.829  \\ 
    \addlinespace[-0.6mm]
    \midrule
    \addlinespace[0mm]
   \multirow{4}{*}{M(French)-to-S(English)}  &\cellcolor{gray!15}\ding{51} 
   &\cellcolor{gray!15}\ding{51}
   &\cellcolor{gray!15}\textbf{6.501} &\cellcolor{gray!15}6.925   \\
    & \ding{51}
    & \ding{55}
    & 1.871 & 12.531  \\ 
    & \ding{55}
    & \ding{51}
    & 6.183 & 6.382  \\ 
    & \ding{55}
    & \ding{55}
    & 6.305 & \textbf{6.198}  \\ 
    \addlinespace[-0.6mm]
    \bottomrule
    \end{tabular}
}

\vspace{-3mm}
\caption{
Ablation on cross-lingual audio-driven generation. ``A2VM'' denotes the Audio-to-Visual Mapper and ``SFP'' denotes the Stochastic Feature Pairing strategy. The notation X(Language)-to-Y(Language) indicates that the model is trained using audio from subject X and tested on subject Y, with the specified language denoting the audio modality.
}
\vspace{-6mm}
\label{tab:cross}
\end{table}

\section{Conclusion}
\label{sec:conclusion}


This paper presents HM-Talker, a novel hybrid motion modeling framework that solves the fundamental challenge in talking head generation: the reconciliation of person-specific anatomical priors with robust cross-identity audio generalization. Our key insight is to hybridize implicit audio features with explicit visual priors. This is realized through two components: a Cross-Modal Mapping Module (CMMM) that constructs a comprehensive, anatomically grounded motion vocabulary by aligning audio prosody with visual articulation, and a Hybrid Motion Modeling Module (HMMM) that uses a Stochastic Feature Pairing (SFP) strategy to dynamically train the model for both personalization and generalization. Extensive experiments confirm that HM-Talker achieves state-of-the-art performance, consistently producing high-fidelity, lip-synchronized, and identity-preserving talking head videos. By unifying previously competing objectives of personalization and generalization, our work establishes a new paradigm for universal facial animation.


\section{Acknowledgement}

This work was supported by the National Natural Science Foundation of China (62501189), the Natural Science Foundation of Heilongjiang Province of China for Excellent Youth Project (YQ2024F006) and Guangdong Basic and Applied Basic Research Foundation (2026A1515010184).
{
    \small
    \bibliographystyle{ieeenat_fullname}
    \bibliography{main}
}


\end{document}